# Structural Design Through Reinforcement Learning

Thomas Rochefort-Beaudoin[a], Aurelian Vadean[a], Niels Aage,[b] Sofiane Achiche[a]

[a]*Department of Mechanical Engineering, Polytechnique Montréal, 2500 Chemin de Polytechnique, Montréal, QC, Canada*

[b]*Department of Civil and Mechanical Engineering, Solid Mechanics, Technical University of Denmark. Koppels Allé, B.404, 2800 Kgs. Lyngby, Denmark*

## Abstract

This paper introduces the Structural Optimization gym (SOgym), a novel open-source Reinforcement Learning (RL) environment designed to advance machine learning in Topology Optimization (TO). SOgym enables RL agents to generate physically viable and structurally robust designs by integrating the physics of TO into the reward function. To enhance scalability, SOgym leverages feature-mapping methods as a mesh-independent interface between the environment and the agent, allowing efficient interaction with the design variables regardless of mesh resolution. Baseline results use a model-free Proximal Policy Optimization agent and a model-based DreamerV3 agent. Three observation space configurations were tested. The *TopOpt* game-inspired configuration, an interactive educational tool that improves students' intuition in designing structures to minimize compliance under volume constraints, performed best in terms of performance and sample efficiency. The 100M parameter version of DreamerV3 produced structures within 54% of the baseline compliance achieved by traditional optimization methods and a 0% disconnection rate, an improvement over supervised learning approaches that often struggle with disconnected load paths. When comparing the learning rates of the agents to those of engineering students from the *TopOpt* game experiment, the DreamerV3-100M model shows a learning rate approximately four orders of magnitude lower, an impressive feat for a policy trained from scratch through trial and error. These results suggest RL's potential to solve continuous TO problems and its capacity to explore and learn from diverse design solutions. SOgym provides a platform for developing RL agents for complex structural design challenges and is publicly available to support further research in the field.

## Introduction

In structural design, the task of Topology Optimization (TO) seeks to distribute a limited amount of material within a given design space under specific design constraints and objectives. Density-based methods [1], and in particular the SIMP method [2], which parameterize the topology of a structure using penalized individual elemental densities in a discretized finite element mesh, became the standard TO framework thanks to their robustness, simplicity of implementation [3], and adaptability to various manufacturing constraints [4] and optimization objective [5]. TO is an iterative procedure where, most often, Finite Element Analysis (FEA) is used to compute the structural response of a given topology under specific boundary conditions. At each iteration, a sensitivity analysis is performed to obtain the gradient of the optimization objective as a function of the design variables.

The application of TO shows potential for reducing the material and energy consumption of large-scale structures such as suspension bridges [6] and entire airplane wings [7], presenting engineers with more sustainable designs in big energy-hungry industries such as transport and infrastructure. Although both FEA and adjoint sensitivity analysis can be performed highly efficiently on massively parallel computing systems, the sparse access to such facilities still curbs the widespread use of the TO method.

Optimizing for nonlinear multi-physics is especially important if the structures are to be manufactured through Additive Manufacturing (AM) which often generates lattice or cellular structures that are more prone to nonlinear deformation such as local buckling and inelastic yield [8]. These physics can be implemented in TO frameworks, but the additional computational cost associated with them escalates substantially, limiting their broader applicability [9].

Motivated to lower the cost barrier of large-scale and high-fidelity TO, researchers have proposed various methods to reduce the number of total iterations required or mitigate the computational expenditure associated with each iteration [10]. Over the past few years, the introduction of machine learning-based methodologies to expedite TO has been noticeably prevalent. Rapid advances in the subfield of deep learning, a branch of machine learning, stimulated a gold rush-like interest in the deployment of its methodologies across a wide range of applications. The parameterization of the TO density field as pixel/voxels has naturally brought many researchers to flock to image-based deep learning methods to try to accelerate the TO process as a low-hanging fruit. Most of these approaches propose a standard supervised learning pipeline of generating optimal topologies, and then training a computer vision model such as a convolutional neural network via a direct regression loss, or in an adversarial context.

The training loss functions used in these supervised learning methods only optimize indirectly for the TO objective and do not consider the actual physics of the TO problem. In previous work, the incompatibility between a regression loss function on the design variables and the TO objective has been demonstrated, showing that a common failure mode of such models is to generate structures with a disconnected load path, and the inability to treat boundary conditions outside the training set [11, 12].

To tackle this issue, deep learning approaches for TO need to be enriched with the underlying physics of the problem [12]. Research has been conducted to reparametrize the density field in TO problems and to integrate the finite element analysis (FEA) operation directly into the model’s loss function, enabling backpropagation through the entire structural analysis [13, 14]. This necessitates the use of sparse solvers in auto-differentiation software, which are currently only experimentally supported in popular deep learning libraries. Consequently, the cost of backpropagation scales not only with the model size but also with the size of the FEA mesh. Surprisingly, all these studies also omit the potential for the structure to be generated in a single forward pass through the model, instead concentrating on achieving an optimal structure by updating the parameters for each new problem through gradient descent on the model’s parameters. Thus, this paradigm presents more as an alternative to typical optimizers, rather than offering a means to leverage past solutions for solving the current problem.

In this paper, the standard compliance minimization objective of TO is framed as a reinforcement learning (RL) task to train a deep learning agent to generate structures as efficiently as possible for various Boundary Conditions (BC). Unlike supervised learning, which is limited by the availability, diversity, and quality of training data, RL operates on the premise of trial-and-error, learning to make decisions based on the maximization of cumulative rewards over time.

Framing TO as a RL task draws inspiration from the interactive *TopOpt* game [15]. In this game, developed at the Technical University of Denmark, players use a user interface to design structures that maximize a score based on compliance minimization while adhering to volume constraints. The game's effectiveness was evaluated in a classroom setting with undergraduate design engineering students. The study aimed to explore the potential of gamifying TO, and results demonstrated a positive correlation between the players' experience and the structural performance of their solutions, suggesting that students' intuition in TO improved with practice.

The literature on using RL for TO is sparse, with only two significant studies addressing this approach. The first study [16] introduced a method using a graph representation of a binary truss topology, where an RL agent was trained to remove unnecessary truss components from a fully connected graph, showing generalization capabilities to various boundary conditions and larger grid resolutions. However, it is worth noting that full FEA is still performed in each iteration and hence, that the speedup is most likely related to an improved design space representation. The second study [17] extended this approach to discretized 2D domains using a fixed grid mesh. Here, an RL agent iteratively removed individual pixels from a fully saturated 2D material mesh until reaching a sub-optimal solution. While these methods share similarities with evolutionary TO algorithms, they are constrained by the limited action space of manipulating individual elements, making it challenging to scale to higher mesh resolutions due to the exponential increase in computational complexity.

This paper introduces the Structural Optimization gym (SOgym), an open-source reinforcement learning (RL) environment. SOgym provides a platform for developing and training RL algorithms specifically for structural design. It recreates the *TopOpt* game's scoring system and user interface, allowing for direct comparisons between the learning rates of engineering students and trained RL agents. Baseline performance results for standard RL algorithms are also included, offering a foundation for future research.

# Method

## The SOgym environment

SOgym is based on the Gym [18] framework and defines the interactions between the agent and the environment. It specifically outlines three key components:

1. The action space: a set of continuous actions that the agent can take, corresponding to the design variables of the topology.
2. The observation space: a representation of the current state of the TO problem.
3. The reward function: feedback provided to the agent based on the performance of its generated structure.

## Action space

This paper uses feature-mapping methods [19] as a more scalable and interpretable interface for RL agents to interact with the material distribution of the TO problem, in comparison to using an elemental density action space similar to density-based TO methods, as implemented in [16] and [17]. This approach offers two key advantages: (1) the number of design variables remains independent of the mesh resolution, enhancing scalability for the action space of the agent, and (2) the design variables exhibit geometric interpretability, which can streamline the post-processing phase of the optimal topology.

SOgym uses a modified implementation of the 188-line Method of Moving Morphable Components (MMC) code [20]. In this approach, the endpoints' coordinates and two thicknesses define a component with a linearly varying thickness, as illustrated in Figure 1 and detailed in Equations 1 through 7. The RL agent sequentially places MMC components in the design space to create the stiffest structure possible within a volume constraint. The maximum number of steps per episode, equivalent to the number of components, is a user-defined variable controlling the structure's complexity. For the baseline experiments in this paper, a maximum of 8 components was used. An episode involves iteratively placing these 8 components, with FEA performed 'per step' or 'per episode' for compliance evaluation, depending on the observation space configurations detailed later in this section. This framework can be scaled to larger models by increasing the number of components, enabling the exploration of more complex structures.

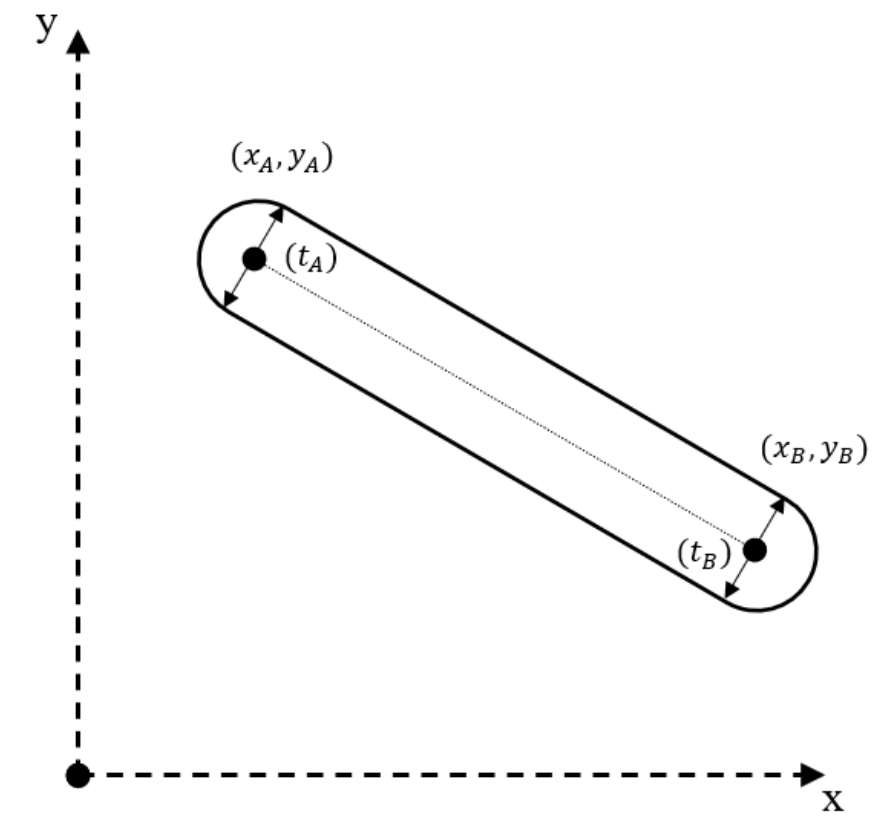

Figure 1: MMC component defined by endpoint coordinates and thicknesses.

$$x_0 = \frac{x_A + x_B}{2}, y_0 = \frac{y_A + y_B}{2} \tag{1}$$

$$L = \sqrt{(x_B - x_A)^2 + (y_B - y_A)^2} \tag{2}$$

$$\theta = arctan2\,(B_y - A_y\,, B_x - A_x) \tag{3}$$

$$x_1 = \cos(\theta)\,(x - x_0) + \sin\,(\theta)(y - y_0) \tag{4}$$

$$y_1 = -\sin(\theta)\,(x - x_0) + \cos(y - y_0) \tag{5}$$

$$t(x_1) = \frac{(t_A + t_B)}{2} + \frac{(t_B - t_A)}{2L}(x_1) \tag{6}$$

$$\Phi(x_1, y_1) = 1 - \left( \left( \frac{x_1}{\frac{L}{2}} \right)^6 + \left( \frac{y_1}{t(x_1)} \right)^6 \right)^{1/6} \tag{7}$$

$x_0$, $y_0$, $L$ and $\theta$ are the coordinates of the centroids, the length, and the orientation of the MMC component, defined by the endpoint coordinates $x_A$, $y_A$, $x_B$ and $y_B$. A linearly varying thickness $t(x_1)$ is defined by interpolating between the two endpoint thicknesses $t_A$ and $t_B$. These geometric variables define the shape of the hyperelliptic function $\Phi(x_1, y_1)$ named the Topology Description Function (TDF). The density distribution on the discretized FEA mesh is obtained using a regularized Heaviside ($H_e$) projection of the TDF, and the ersatz material model is used to define the elemental densities $\rho_e$ and Young's modulus $E_e$. The projection scheme and the ersatz model are described in equations 8 through 10. A regularization parameter $\epsilon$ is used with a value of 1e-2, and a small positive value $\alpha$ of 1e-9 helps avoid numerical singularities with the stiffness value of void elements.

$$H_e(\Phi_i^e) = \begin{cases} 1, \quad if\ \Phi_i^e > \epsilon \\ \frac{3(1-\alpha)}{4}\left(\frac{x}{\epsilon} - \frac{x^3}{3\epsilon^3}\right) + \left(\frac{1+\alpha}{2}\right), \quad if\ |\Phi_i^e| \le \epsilon \\ \alpha \quad otherwise \end{cases} \tag{8}$$

$$\rho_e = \frac{1}{4}\sum_{i=1}^{4} H_e(\Phi_i^e) \tag{9}$$

$$E_e = \rho_e E \tag{10}$$

The continuous action space of the agent is a vector of 6 elements with values normalized between [-1,1] corresponding to the two endpoint coordinates as well as the two endpoint thicknesses. The actions are scaled to the design space using min/max values defined in Table 1.

Table 1: Minimum and maximum boundaries of the design variables

| Design variable | Definition | Minimum value | Maximum value |
|---|---|---|---|
| $x_a$ | X-coordinate of endpoint a | 0.0 | Width of domain |
| $x_b$ | X-coordinate of endpoint b | 0.0 | Width of domain |
| $y_a$ | y-coordinate of endpoint a | 0.0 | Height of domain |
| $y_b$ | y-coordinate of endpoint a | 0.0 | Height of domain |
| $t_a$ | Thickness of endpoint a | 0.01 | 5% min(h,w) |
| $t_b$ | Thickness of endpoint b | 0.01 | 5% min(h,w) |

## Boundary conditions

Instead of relying on the typical train-test set of Boundary Conditions (BC) common in machine learning TO literature, SOgym introduces a large, random distribution of BC at the beginning of each episode. This distribution is extensive enough to make encountering every possible combination impractical within a reasonable training timeframe. As a result, the agent consistently faces new and varied challenges, encouraging the development of generalized problem-solving skills instead of overfitting to a limited set of problems.

A 2D rectangular domain was used where support and load conditions, desired volume fractions, and domain dimensions were varied to generate multiple unique parameter combinations, bringing sufficient diversity during training. The distribution of parameters, which were all sampled from uniform distributions with ranges specified in the third column, is illustrated in Figure 2 and detailed in Table 2. The support condition is placed randomly on one of the four boundaries without spanning multiple boundaries, covering only a fraction of the chosen boundary at a time. The point load is then placed randomly on the opposite boundary, with a random orientation and a unit load magnitude. A coarse mesh resolution of 50 elements per unit of height and width was set for the FEA to limit the computational cost of running the environment.

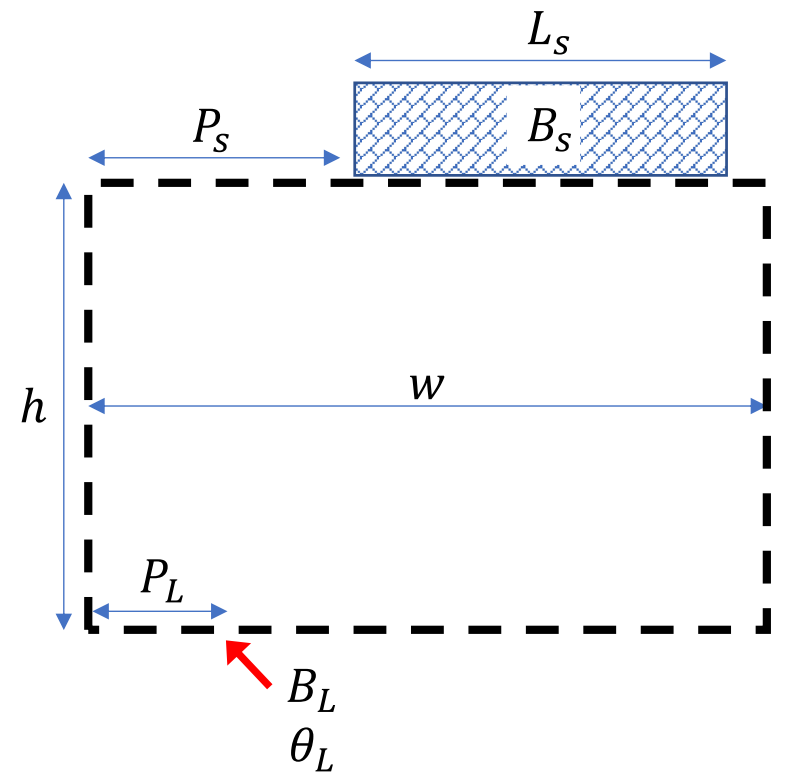

Figure 2: Boundary conditions parameters defining the random distribution

Table 2: SOgym environment parameters

| Parameter | Description | Possible values |
|---|---|---|
| $L_s$ | Length of support | [25%, 100%] of boundary |
| $B_s$ | Boundary of support | Left, Right, Top, Bottom |
| $P_s$ | Position of support | [0%, 100%-$L_s$] |
| $B_L$ | Boundary of the load | Opposite of $B_s$ |
| $P_L$ | Position of the point load | [0%, 100%] of boundary |
| $\theta_L$ | Orientation of load | [0°, 360°] |
| h, w | Height, Width of domain | [1.0, 2.0] |
| V | Desired volume fraction | [20%, 40%] |

## Observation space

SOgym offers three observation space configurations: *Vector*, *Image*, and a simplified imitation of the *TopOpt* game interface, as detailed in Table 3.

1. *Vector* configuration: It includes a boundary conditions description vector ($\beta$), based on [11] and detailed in Table 4. This vector encapsulates all the parameters defining the current TO problem. Although our baseline experiments use a single point load and a fully supported support type, the $\beta$ vector was designed to handle up to five-point loads and multiple support types. Additionally, this configuration includes:
    - A fraction indicating the number of remaining steps in the episode,
    - A vector containing the values of all the current design variables,
    - A fraction representing the current volume.
2. *Image* configuration: It adds a low-resolution image of the current design, illustrating support and load arrows.
3. *TopOpt* game configuration: It adds the current score and an image of the normalized strain energy in a jet color scheme, only shown if the load path is connected. Figure 3 illustrates this configuration at the end of an episode with random actions and a connected final structure.

The *Vector* configuration is more compact, while the *Image* and *TopOpt* game configurations offer richer observations at the expense of higher computational costs. The *TopOpt* game configuration demands solving the displacement field at every iteration, unlike the others, which only do so at the episode's end. This complexity enriches the RL agent's training data but slows down training due to extra computations and solver calls. The trade-off between the richness of the observation space and the computational efficiency of the environment is an important consideration that will be further explored and analyzed in the results section of this paper.

Table 3: Supported observation space configurations of the SOgym environment.

| Name | Description | Shape | Observation Space Configurations *Vector* | *Image* | *TopOpt Game* |
|---|---|---|---|---|---|
| $\beta$ | Boundary conditions description vector | (27,1) | ✓ | ✓ | ✓ |
| # steps left | Number of steps (components) left for the problem | (1,1) | ✓ | ✓ | ✓ |
| Design variables | Vector with all the design variables of the current design | $(6 \cdot t_{max}, 1)$ | ✓ | ✓ | ✓ |
| Volume | Current volume at iteration t | (1,1) | ✓ | ✓ | ✓ |
| Image | Low-resolution image of the current design with distinct color MMC components | (3,64,64) | | ✓ | ✓ |
| Strain energy | The logarithm of the current strain energy plotted using the jet color scheme | (3,64,64) | | | ✓ |
| Score | Current score at iteration t | (1,1) | | | ✓ |

Table 4: Composition of the $\beta$ description vector.

| Component | Sub-Component | Details | Index Range |
|---|---|---|---|
| Support conditions | Selected Boundary | Boundary selected for support | 1 |
| | Support Type | Type of support applied | 2 |
| | Boundary Length | Length of the boundary | 3 |
| | Boundary Position | Position of the boundary | 4 |
| Load conditions | Load Positions | Position of loads | 5-9 |
| | Load Orientations | Orientation of loads | 10-14 |
| | Magnitudes x | x-component of point load magnitudes | 15-19 |
| | Magnitudes y | y-component of point loads magnitudes | 20-24 |
| Volume fraction | - | Volume fraction constraint | 25 |
| Domain shape | Width | Domain width (x-dimension) | 26 |
| | Height | Domain height (y-dimension) | 27 |

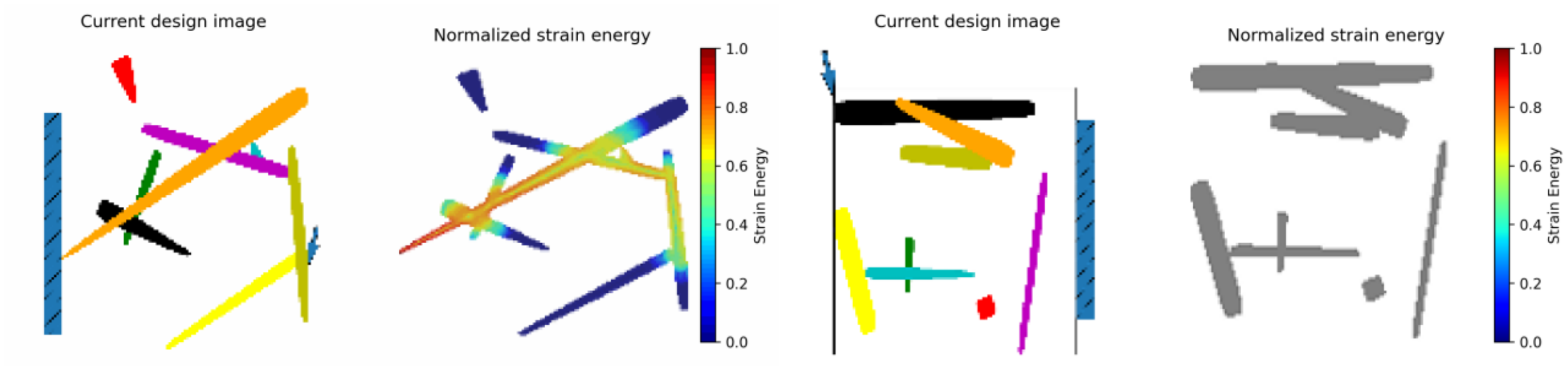

| | |
|---|---|
| **Design Variables** | [0.33, 0.08, 0.15, 0.01, 0.01, 0.16, 0.16, 0.25, 0.08, 0.01, 0.01, -0.63, 0.13, 0.44, 0.04, 0.02, 0.01, -0.37, 0.37, 0.32, 0.01, 0.02, 0.00, -0.81, 0.30, 0.32, 0.15, 0.01, 0.01, 0.92, 0.46, 0.19, 0.14, 0.01, 0.01, 0.54, 0.16, 0.20, 0.10, 0.00, 0.02, 0.87, 0.24, 0.29, 0.27, 0.00, 0.02, 0.16] |
| **Steps Left** | 0.00 |
| **Volume** | 0.15 |
| **Beta** | [0.00, 0.00, 0.70, 0.06, 0.28, 0.71, -0.26, -0.97, 0.00, 0.00, 0.00, 0.00, 0.00, 0.00, 0.00, 0.00, 0.00, 0.00, 0.00, 0.00, 0.00, 0.00, 0.00, 0.00, 0.32, 1.50, 1.60] |
| **Score** | 0.08 |

a)

| | |
|---|---|
| **Design Variables** | [0.04, 0.11, 0.08, 0.02, 0.02, 0.60, 0.13, 0.12, 0.04, 0.00, 0.01, 0.49, 0.26, 0.06, 0.01, 0.01, 0.01, 0.12, 0.15, 0.11, 0.09, 0.01, 0.00, -0.00, 0.34, 0.12, 0.12, 0.01, 0.01, 0.44, 0.19, 0.26, 0.08, 0.01, 0.01, -0.02, 0.17, 0.32, 0.17, 0.01, 0.02, -0.99, 0.21, 0.29, 0.11, 0.01, 0.02, -0.12] |
| **Steps Left** | 0.00 |
| **Volume** | 0.19 |
| **Beta** | [0.25, 0.00, 0.75, 0.13, 0.96, 0.79, 0.26, -0.97, 0.00, 0.00, 0.00, 0.00, 0.00, 0.00, 0.00, 0.00, 0.00, 0.00, 0.00, 0.00, 0.00, 0.00, 0.00, 0.00, 0.32, 1.30, 1.10] |
| **Score** | 0.00 |

b)

Figure 3:Illustration of the TopOpt observation space. a) Example of a random episode leading to a connected final structure. b) Example of a random episode leading to a disconnected structure where the strain energy is not shown.

## Reward function

Like the *TopOpt* game, a sparse reward is given only at the final timestep ($t_{max}$) and corresponds to the inverse of the compliance with a hard constraint on volume and load-path connectivity. The function sets the reward to 0 if the structure's final volume exceeds the defined constraint or if the load and support are disconnected. Since the magnitude of the compliance can vary widely depending on the boundary conditions and the generated structure, SOgym uses the natural logarithm of the compliance to squash the reward signal. This reward-squashing strategy, shown in Figure 4, has been demonstrated to enable RL algorithms to generalize better across diverse environments [21]. Equation 11 shows the SOgym reward given at $t_{max}$ where $V^*$ is the volume constraint, $V_{t_{max}}$ is the volume at the final timestep and $K_{t_{max}}$ is a boolean variable that is equal to 1 if the load-path of the final structure is connected.

$$r_{t_{max}} = \begin{cases} \dfrac{1}{\log_e(compliance_{t_{max}})}, & \Leftrightarrow \left(V_{t_{max}} \leq V^* \; and \; K_{t_{max}} = 1\right) \\ 0\,, & elsewhere \end{cases} \quad (11)$$

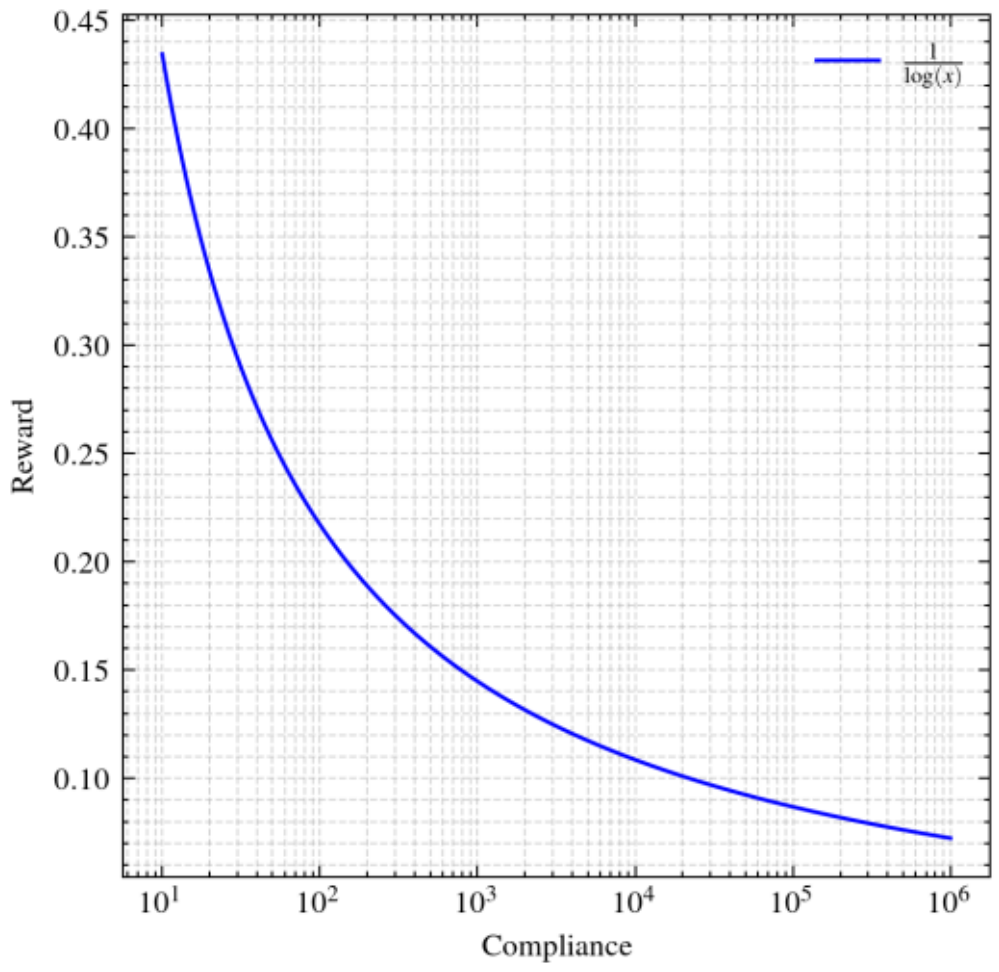

Figure 4: Squashed SOgym reward using the inverse of the natural logarithm of the compliance.

## RL algorithms

Two RL algorithms, one model-free and one model-based, are trained and tested to establish an initial performance baseline for SOgym. Proximal Policy Optimization (PPO) [22] is a popular model-free, on-policy algorithm known for its simplicity, robustness, and successful applications in various domains, including chip design [23], defeating world champions in complex esports [24], and human alignment of large language models [25]. DreamerV3 [21], a model-based RL algorithm, is chosen for its innovative blend of policy optimization and world model learning. It utilizes a learned world model to create imagined trajectories, predicting environmental responses to specific actions. The world model consists of three key components: a representation model that compresses observations into latent states, a transition model that predicts the next latent state based on the current state and action, and a reward model that estimates rewards from these latent states. This approach enables DreamerV3 to simulate future outcomes, plan strategically, and explore the environment more efficiently, resulting in improved sample efficiency in most RL benchmarks.

## Model architecture

Two versions of DreamerV3 were trained: the smallest model, with 12 million parameters, and a larger model with 100 million parameters. The 12M parameter model was chosen because it was the smallest tested in [21], providing a baseline for comparison. The 100M parameter model was selected as it represents a middle-ground model, balancing complexity and performance. Details about the model architectures and default hyperparameters used for the baseline results of this paper are provided in [21]. For the PPO algorithm, the IMPALA CNN network architecture [26] was chosen due to its effective balance of performance and computational cost [27]. The CNNs used as a feature extractor for the image inputs are detailed in Figure 5. Vector inputs are processed through a single fully connected layer to extract features. These feature embeddings are concatenated and passed to an LSTM layer, which then feeds into both the policy and value networks. The architecture illustrated in Figure 5 is repeated for the policy and the value networks, as they do not share feature extractors. The PPO hyperparameters used in [21] are used as they were selected based on recommendations for on-policy RL algorithms [28].

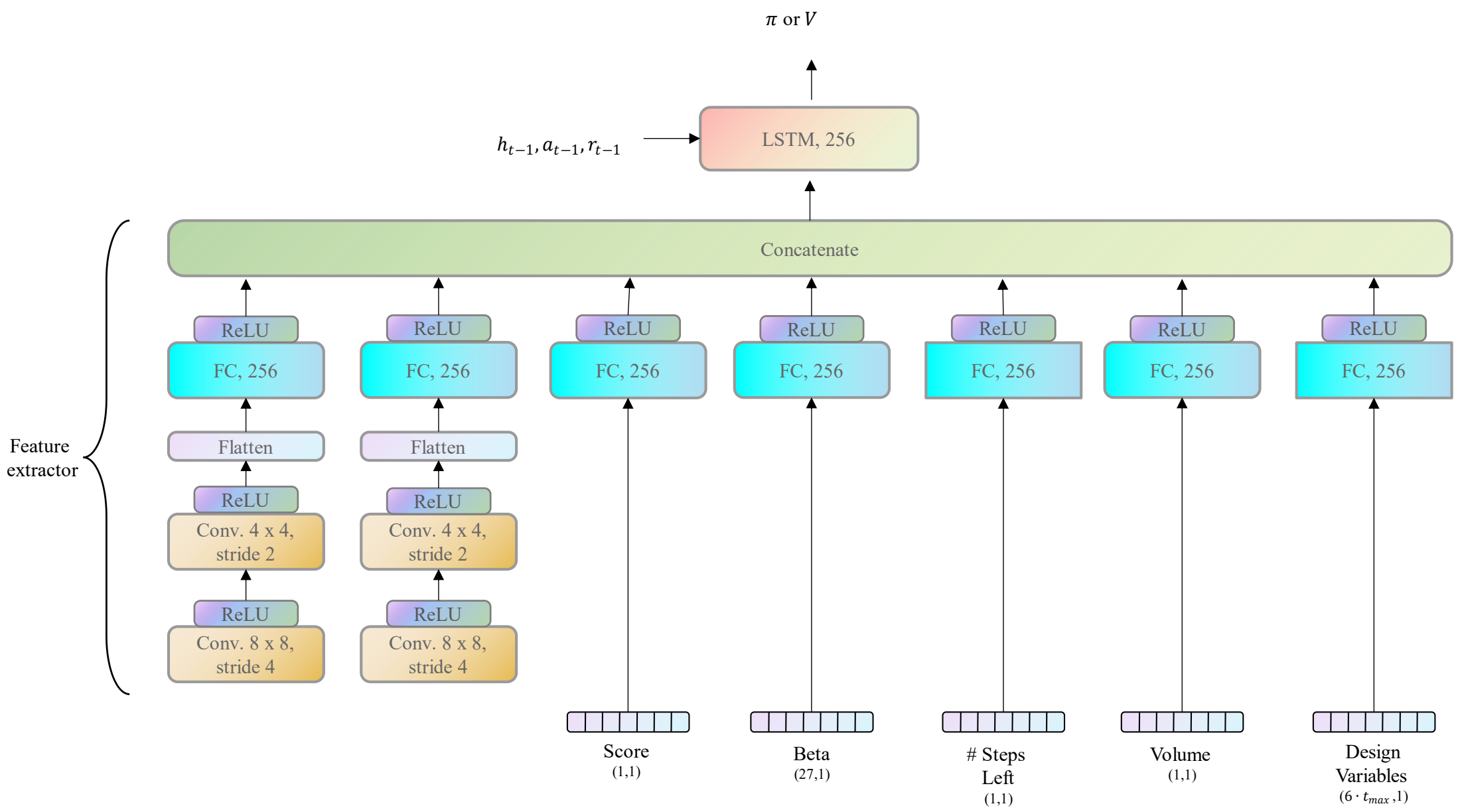

Figure 5: Actor-critic IMPALA-inspired architecture of the PPO model.

# Results

To monitor the models' performance during training, a fixed set of 10 problems was randomly sampled from the BC distribution to serve as evaluation benchmarks. This number was kept small to limit the impact on the training duration. At regular intervals during training, the agent evaluates its performance by greedily sampling actions across these 10 problems. The mean evaluation reward from these runs is used to assess the performance across the three observation space configurations and the two baseline algorithms. This evaluation reward was also used to calculate a learning rate on the same basis as the in-class experiment from the *TopOpt* game.

All training runs were conducted on a single compute node equipped with an Nvidia A100 GPU and a 48-core CPU. The requirement for FEA to evaluate rewards makes the environment CPU-bound. To mitigate this bottleneck 48 parallel copies of the environment and policy are run simultaneously to aggregate experiences between gradient updates efficiently.

## Evaluation of performance across different observation spaces

The impact of the observation space was first tested on the Dreamerv3-12m model since it was the smallest model tested in terms of the number of trainable parameters. Due to computational constraints, the first training runs with DreamerV3-12M are limited to 48 hours. As illustrated in Figure 6a) and Figure 6b), the *Vector* observation space configuration completed just under 6M training episodes and reached a best evaluation reward of 0.27 for an average of 279 Frames Per Second (FPS), while the *Image* configuration completed around 3.75M episodes with a best evaluation reward of 0.28 with an average FPS of 173. The *TopOpt* game configuration showed a much slower FPS of 112, but it managed to reach a higher maximum evaluation reward of 0.29 in just under 2.5M episodes.

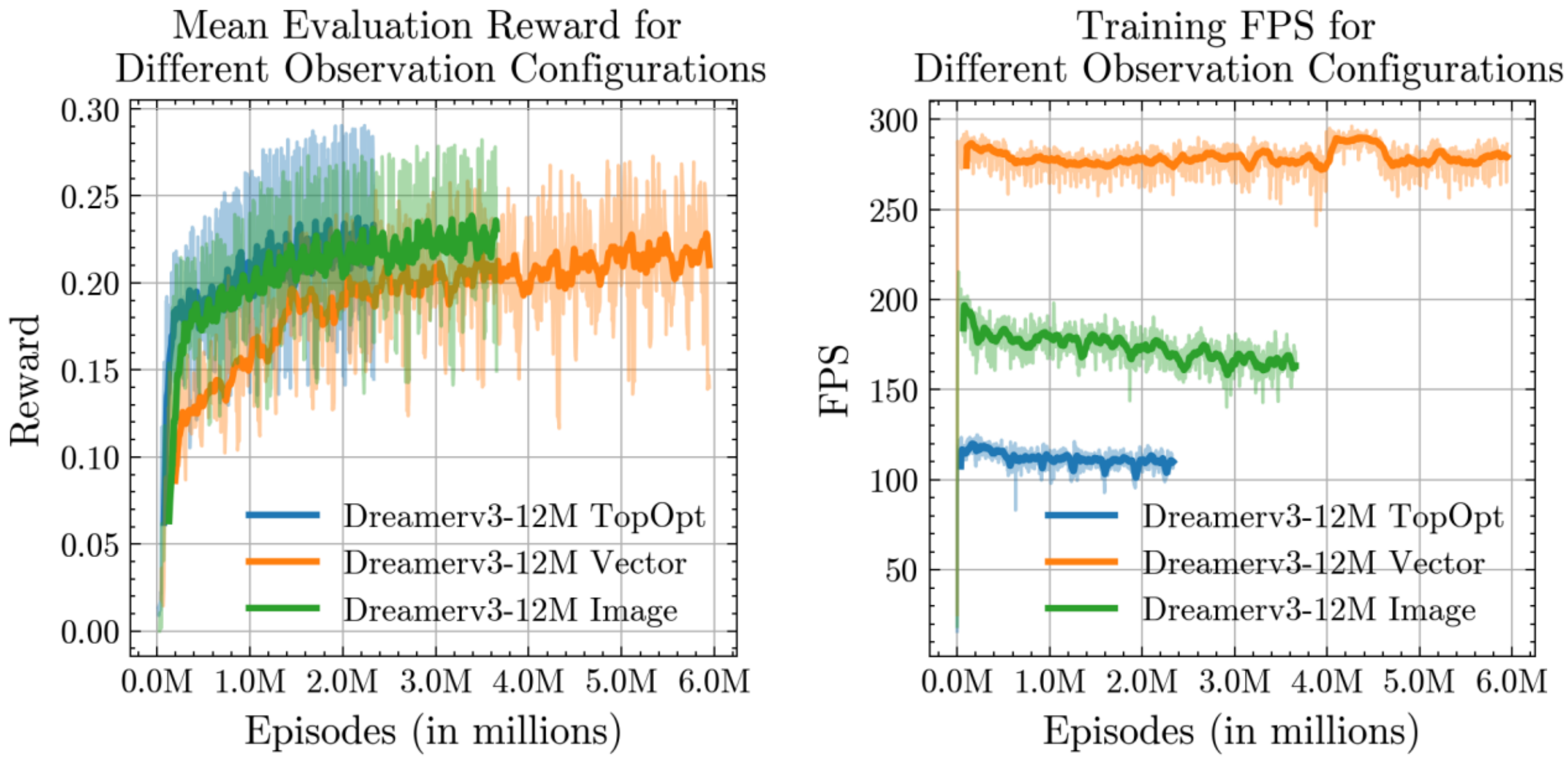

a) b)

Figure 6: Performance comparison of different observation spaces for DreamerV3-12M model. a) Mean evaluation reward: *TopOpt* (blue) achieves the highest reward of 0.29 in around 2.3M episodes. *Image* (green) achieves a maximum reward of 0.28 in around 3.75M episodes, while *Vector* (orange) reaches 0.27 after nearly 6M episodes. b) Training Frames Per Second (FPS): *Vector* (orange) maintains the highest FPS at 279, followed by *Image* (green) at 173 FPS. *TopOpt* (blue) has the lowest FPS at 112.

## Dreamerv3 vs PPO

The *TopOpt* observation space was used as a benchmark to compare the performance of the DreamerV3-12M, DreamerV3-100M, and PPO agents. To investigate whether increasing the training budget could enhance performance, as the DreamerV3-12m evaluation reward in Figure 6a) had not plateaued after the initial 48-hour time budget, an additional 96 hours of training time was allocated, bringing the total training time for each algorithm to 6 days. Figure 7 illustrates the evaluation rewards throughout the training period for the three algorithms. Both versions of DreamerV3 performed better than the PPO agent, with maximum evaluation rewards reaching 0.30 for the 100M parameter version, and 0.29 for the 12M parameter version. They showed a rapid increase in evaluation reward and a slower rate of learning after about 1M episodes. The PPO agent showed a slower rate of increase in evaluation rewards, reaching a maximum value of 0.19 before suffering from a significant episode of instability, causing the evaluation reward to drop by more than 25%. However, this performance drop was mitigated by saving the best-performing version of the model during training. This ensured that even if the model's performance declined later, the best result achieved up to that point was preserved. This instability coincided with a spike in Kullback-Leibler (KL) divergence, as can be observed in Figure 8, which measures the divergence between the new and old policy distributions, providing insight into the algorithm's learning stability. An increase in KL divergence indicates potential training instability.

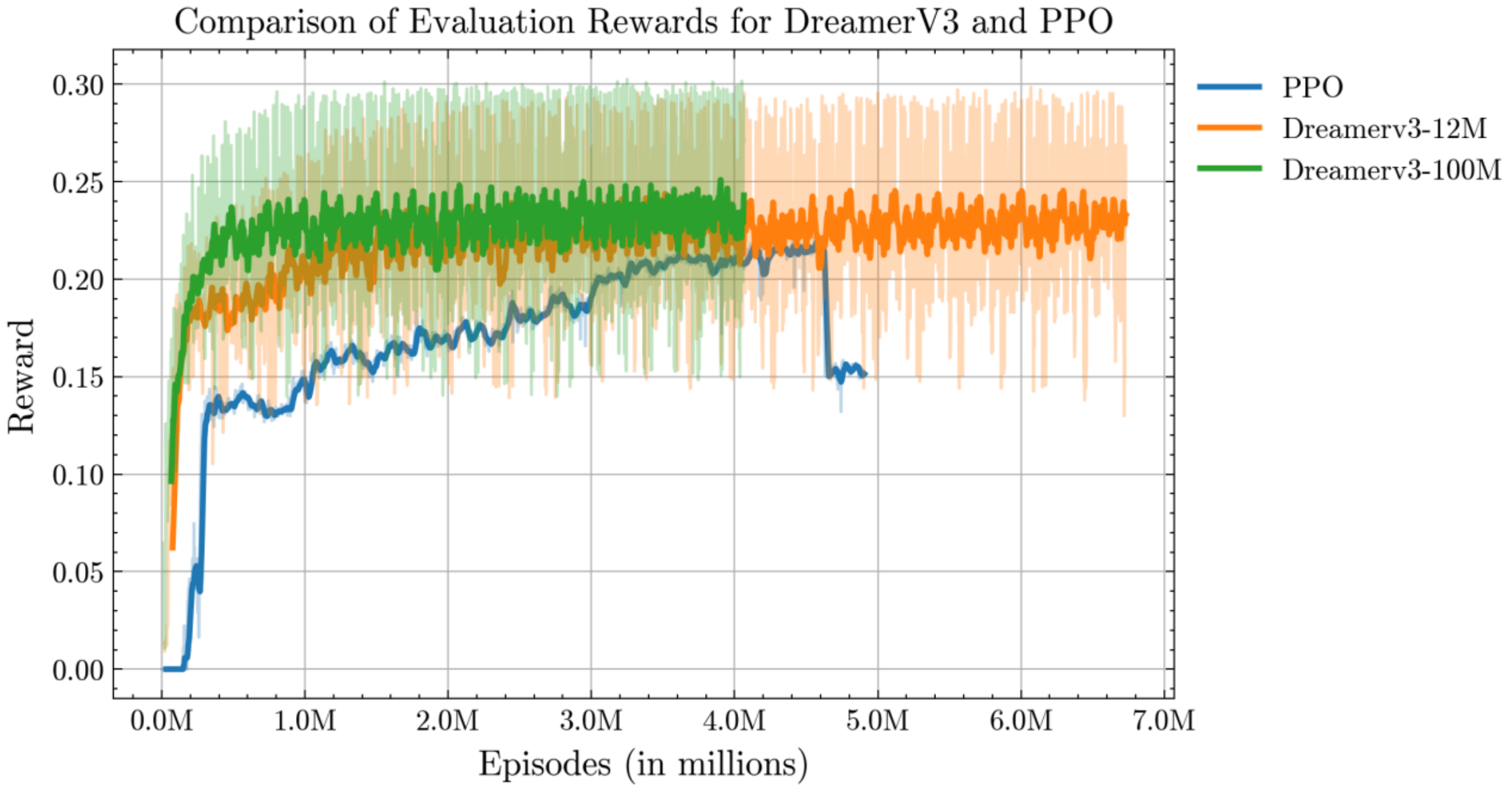

Figure 7: Comparison of evaluation rewards for DreamerV3-12M (orange), DreamerV3-100M (green), and PPO (blue). PPO reaches the highest evaluation performance while the two DreamerV3 show the fastest pace of improvement early in training and then rapidly top off without improving much.

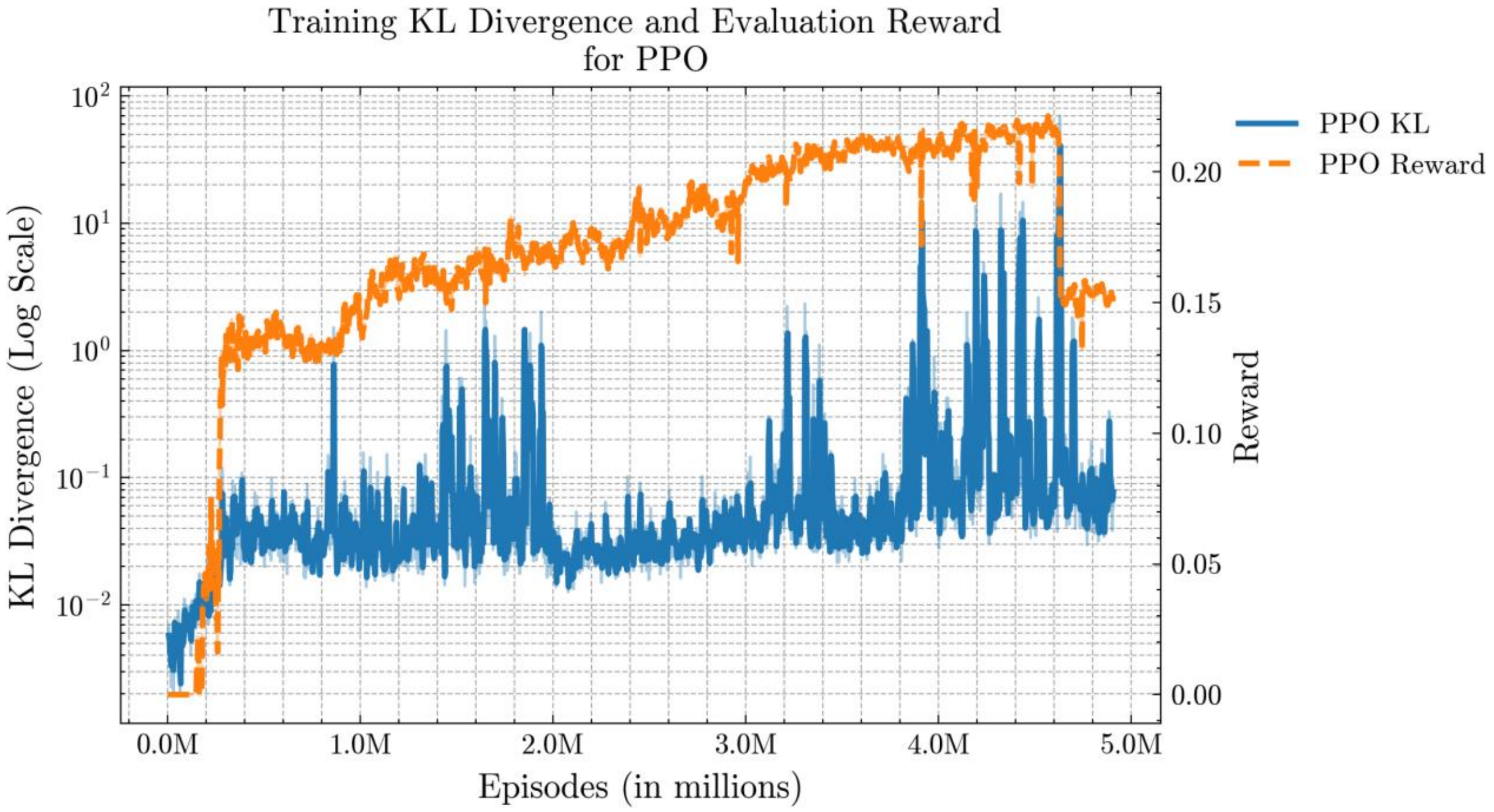

Figure 8: PPO Training Dynamics: KL Divergence and Evaluation Reward. This figure shows the Kullback-Leibler (KL) divergence in logarithmic scale (in blue) and evaluation reward (in orange) during PPO training. The KL divergence generally rises, indicating learning instability, with a sharp spike around the 4.6 million episodes that aligns with a significant drop in evaluation performance, highlighting instability in the training process.

## Learning rate comparison

A linear regression was fitted to the points on the training curve up to the moment of best evaluation performance for each algorithm. To match the scoring method of the *TopOpt* game paper [15], the evaluation reward was converted to inverse compliance and normalized using the maximum observed evaluation reward across the three RL training runs. The origin of the linear regression was forced to zero since the three agents started from scratch. Figure 9 illustrates the best regression lines for all three agents alongside the linear regression derived from the in-class experiment. The DreamerV3-100M model demonstrated the fastest learning rate among the three RL agents, with a value of 1.78e-07 per episode. This rate is nearly four orders of magnitude lower than the 2.40e-03 learning rate observed among the engineering students in the in-class experiment. The DreamerV3-12M agent followed with a learning rate of 9.13e-08, and PPO had the lowest learning rate of 5.46e-08.

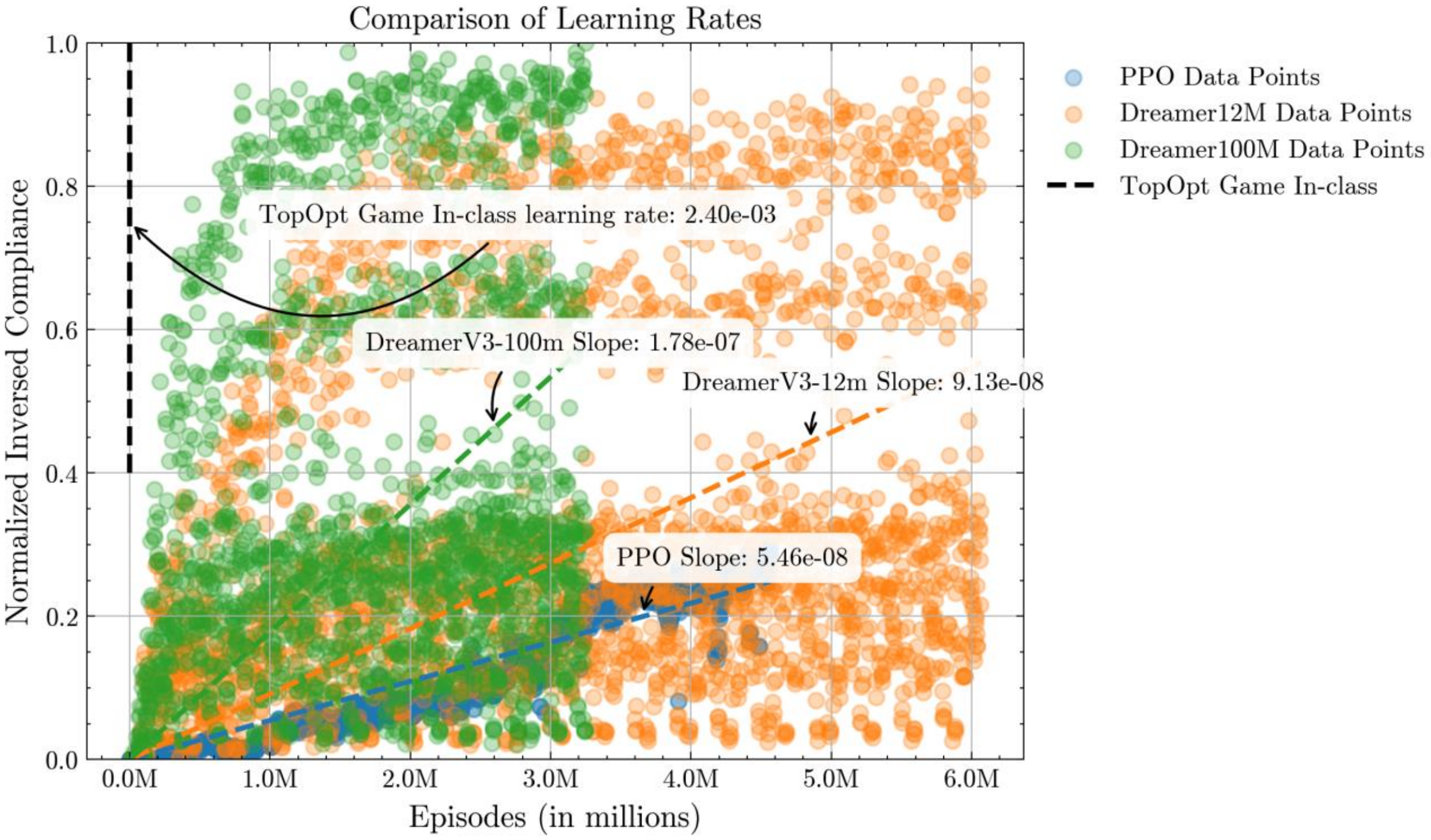

Figure 9: Normalized inverse compliance over episodes for PPO (blue), DreamerV3-12M (orange), and DreamerV3-100M (green). The DreamerV3-100M demonstrated the highest learning rate of 1.78 -07, just under 4 orders of magnitude smaller than then the learning rate of the *TopOpt* game in-class experiment (2.40e-03, black vertical line).

## Structural performance comparison with baseline

The performance of the generated structures from the 3 RL agents was tested using 1000 random boundary conditions problem sampled uniformly from the SOgym distribution. A baseline structure was established with an optimal design obtained via conventional optimization. A hybrid optimizer, integrating both MMA and GCMMA [29], was utilized because of its superior convergence speed on similar problems within the MMC framework [30]. The 188-line MMC implementation [20] in Python was used and the parameters defining the optimization process are presented in Appendix A. The RL agents' generated test structures, presented in Table 5, were evaluated using the following metrics:

- **Median compliance deviation:** The median difference in compliance compared to the optimal designs. The median value is used due to the possibility of having very large outliers in structures with a disconnected load path.
- **Disconnection rate**: The proportion of generated structures with disconnected load paths.
- **Mean volume deviation:** The mean deviation in volume between the generated structures and the desired volume fraction constraint.

Table 5: Performance of the RL-generated structures compared to the corresponding baseline design obtained via conventional MMC optimization.

| Algorithm | Median Compliance Δ (%) | Disconnection Rate (%) | Mean Volume Δ (%) |
|---|---|---|---|
| **PPO** | +219.47% | 0.1% | -29.30% |
| **DreamerV3 12M** | +56.19% | 0.0% | -20.14% |
| **DreamerV3 100M** | +53.96% | 0.0% | -19.45% |

In evaluating the 1,000 test samples, the PPO agent's performance significantly lagged the TO baseline, with its generated structures exhibiting a median compliance over 219% higher than the optimal designs, indicating substantially lower stiffness. Additionally, the PPO agent's designs averaged 29.30% below the constraint, suggesting an inefficient use of the allowed volume. Despite these shortcomings, only one structure from PPO displayed a disconnected load path. In contrast, the DreamerV3 agents demonstrated considerably better performance. The 12M parameter DreamerV3 model produced structures with a median compliance of 56.16% higher than the baseline, while the 100M parameter model showed a slightly lower median compliance difference of 53.96%. Furthermore, the 100M model approached the volume constraint more closely, with an average deviation of -19.45%, compared to -20.14% for the 12M model. Critically, both DreamerV3 agents achieved a 0% disconnection rate, ensuring fully connected load paths in all generated structures. Figure 10 displays 5 randomly selected samples from the 1000 test problems, showcasing the structures generated by the three RL agents compared to the baseline. These examples illustrate the general trends observed in Figure 7 and Table 5, with the DreamerV3-100M agent typically producing the stiffest structures.

Table 6 compares the total processing time required to generate structures for the 1000 test samples using the baseline conventional optimization and the trained RL algorithms. To account for parallel processing, the compute time across the different methods is normalized by reporting the minutes used per CPU worker. The inference times for the trained RL models were 18 minutes for PPO and 28 minutes for both DreamerV3-12M and DreamerV3-100M to process all 1000 problems. In contrast, the conventional iterative optimizer required a total of 17,792 CPU-worker minutes. While the RL models demonstrate impressive inference speeds, it is important to factor in the substantial computational resources consumed during their training phases. By simulating early stopping and considering only the training episodes up to the maximum evaluation reward (the point where the best model was saved), it was observed that the fastest RL algorithm, DreamerV3-100M, consumed 331,526 CPU-worker minutes over 6 days of training, nearly 19 times more than the conventional optimization process.

Table 6: Processing time comparison between conventional optimization and RL algorithms

| Method | Processing time (CPU-Worker minutes)<br>Inference only<br>(1000 test samples) | Training<br>(total) | Training<br>(with simulated early stopping) |
|---|---|---|---|
| Baseline optimizer | 17 792 | - | - |
| PPO | 18 | 414 720 | 386 021 |
| DreamerV3-12M | 28 | 414 720 | 373 911 |
| DreamerV3-100M | 28 | 414 720 | 331 526 |

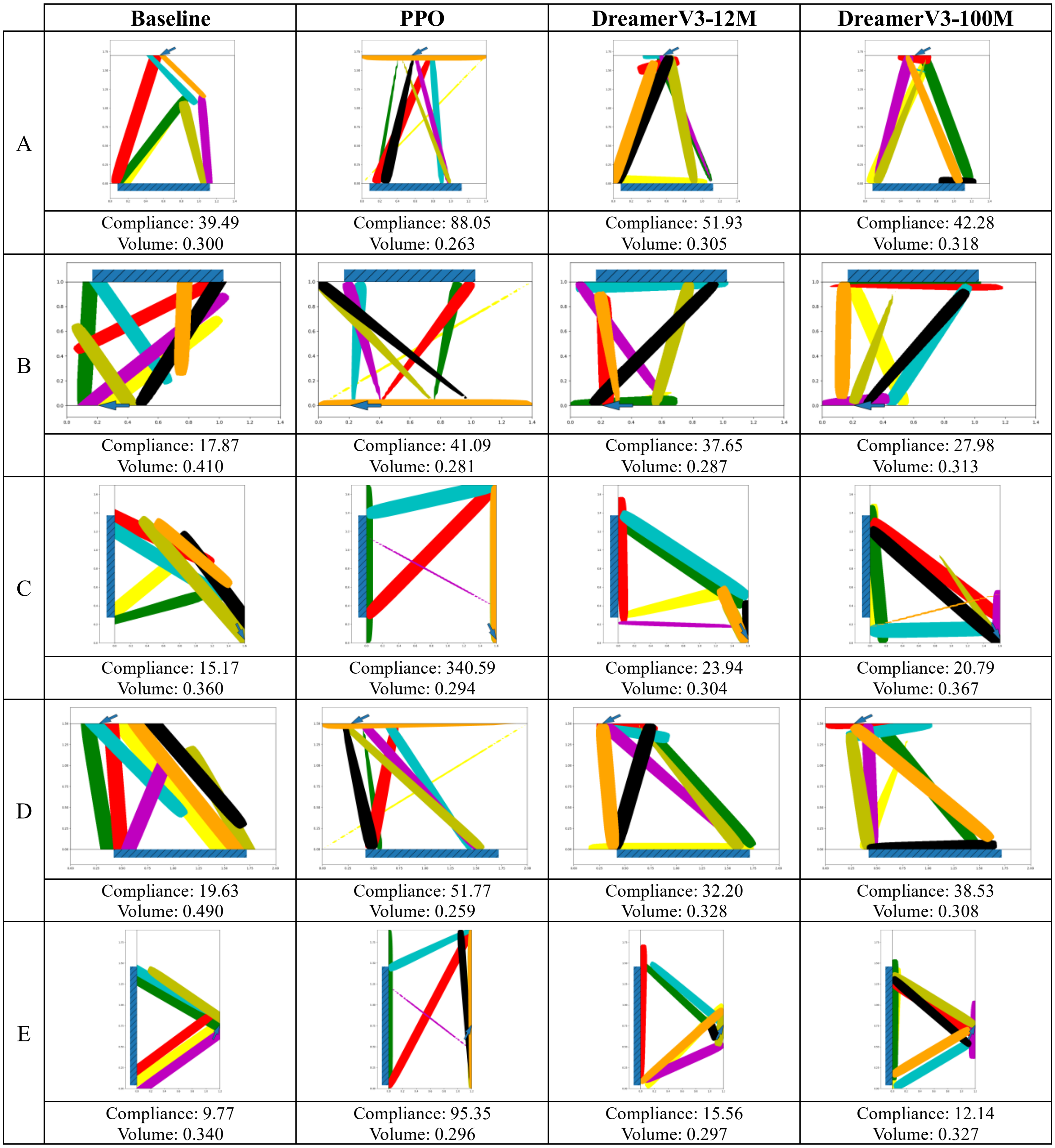

Figure 10: Comparison of generated structures from the RL agents and the corresponding baseline. The samples were randomly chosen from the 1000 test samples.

# Discussion

The baseline results have shown that the SOgym environment is effective for the development of RL agents for the task of TO. Although the trained baseline agents showed worse structural performance when compared to the baseline, it is noticeable that they have achieved this level of performance from a random policy, discovering stiffening patterns and load-path connectivity through pure trial-and-error exploration.

## Algorithm performance and structural quality

A common pattern across the RL-generated designs that can be observed in Figure 10 is the strategic placement of MMC components along the boundary where the support and load are positioned. This behavior is likely influenced by the load-path connectivity requirement in the reward function. During the early stages of training, RL agents seem to adopt a straightforward approach to ensure connectivity by spanning the entire support and load boundaries with components. This can be seen prominently in the PPO-generated samples of Figure 10, where an MMC component consistently covers the boundary area where the load is applied. The DreamerV3 agents develop a more refined approach for the load boundary as they are more accurate in the placement of the components to connect the load, resulting in more efficient designs. They still exhibit components spanning the support boundary.

Although the implementation of the environment in this paper is made to stick as close as possible to the parameters of the *TopOpt* game experiment, soft design constraints like the one described in Equation 12 could provide a denser reward scheme and provide more information to the agent which could help address the behavior of covering the entire support and load boundary with material. Also, the usefulness of using the load path connectivity boolean should be investigated as this also has the impact of sparsifying the feedback from the environment early on during training.

$$r_{t_{max}} = \frac{1}{\log_e(compliance_{t_{max}})} \cdot \left(1 - \left|V_{t_{max}} - V^*\right|\right)^2 \tag{12}$$

## Breakeven threshold

When looking at the breakeven threshold, an economic measure proposed in [31] which evaluates the number of problems that the model needs to generalize to compensate for its training cost, the DreamerV3-100M agent exhibits a breakeven threshold of about 19 000. This is derived from the fact that the total processing time for the RL agent, including training, is approximately 19 times longer than that of the baseline optimization method for generating 1000 structures. However, the method shows good potential to be useful in practice if the trained agent can get to a structural performance level that is more competitive with the conventional optimization process.

## Richer observation space brings training efficiency and performance

The evaluation across different observation space configurations revealed the trade-offs between computational cost and the richness of the observation space. The *TopOpt* game configuration, which provides the most detailed observations including the load path and strain energy visuals, facilitated the highest reward among the tested configurations at the expense of slower training speeds due to the increased computational demands of computing the displacement field at each iteration and processing the additional image observations. However, the higher evaluation reward was reached in a smaller number of episodes, showing that the higher training efficiency per environment interaction offsets the slower FPS.

## Model-based DreamerV3 outperforms model-free PPO

The sample efficiency and performance demonstrated by DreamerV3 highlight the benefits of using a world model for planning. In compute-intensive environments like SOgym, where interacting with the environment is inherently slow, this efficiency is particularly crucial. The ability to predict the future state of strain energy based on material placement is intuitively fundamental for designing efficient structures. By understanding how material changes impact strain energy distribution, the agent can make better decisions. Accurately predicting the outcome of placing components and thus altering the strain field is essential for optimizing performance in complex design tasks. This highlights the importance of world models in applying RL to topology optimization.

Increasing the training budget from 48 hours to 6 days did not improve significantly the performance of the DreamerV3-12M model. Both DreamerV3-12M and DreamerV3-100M exhibited plateaus in their evaluation curves, which might be due to insufficient exploration caused by a lower entropy coefficient (3e-4) compared to PPO's higher entropy coefficient (1e-2). This lower exploration could limit the models' ability to discover more optimal design solutions. Further optimization of hyperparameters could perhaps enhance both sample efficiency and structural performance. Deep reinforcement learning methods are susceptible to these parameters, and careful tuning has been shown to impact performance in various applications dramatically [32].

## A “physical” RL training objective solves the disconnection problem

A noteworthy result is the 0% disconnection rate of the structures generated by the RL agents. This outcome suggests the effectiveness of embedding the physics of the TO problem directly into the training objective. This helps address the common limitation in direct-design TO approaches, where disconnection rates can soar as high as 60% in out-of-distribution test cases [11]. Framing the TO objective as an RL problem also has the added benefit of simplifying the implementation by avoiding the need for backpropagation through sparse solvers. The RL approach relies solely on the forward pass through the FEA and estimates gradients from the rewards gathered over trajectories. This avoids integrating sparse solvers with automatic differentiation software, though it does result in a more sample-inefficient learning process due to its reliance on the less direct form of feedback of trial and error.

## Comparing learning rates with engineering students

The comparison of learning rates between the RL agents and the engineering students in the *TopOpt* game experiment reveals a notable difference. The DreamerV3-100M model, the best-performing RL agent, showed a learning rate about four orders of magnitude lower than that of the students. This significant difference highlights the advantage humans have in applying their intuition and prior knowledge to new and complex tasks. Even though the RL agent’s learning rate is much slower, it is still impressive considering that, for an equal comparison, one would need to compensate for the inherent visual and cognitive abilities that the students bring to the task.

During the in-class experiment, the engineering students received a brief 5-minute introduction explaining the basic theory of TO for compliance minimization. This session provided them with crucial insights into the importance of material distribution, explaining that areas depicted in red on the strain-energy image indicated regions needing more material and that an optimal design should aim for a uniform distribution of strain energy. To mirror this added context, future research could study the integration of a dual-objective reward function that incorporates the notion of uniform strain energy distribution into the agent's objective. This concept, as described in Equation 13, suggests scaling the minimum compliance objective by the normalized standard deviation of the strain energy. This reward function could encourage RL agents to produce designs that are both stiff and demonstrate a balanced strain energy distribution across the design domain, thereby providing richer feedback to the agent.

$$r_t = \frac{1}{\log_e(compliance)} \cdot (1 - \frac{\sigma(W)}{\sigma(W) + \mu(W)}) \quad (13)$$

In addition to refining the reward function, another proposed idea to enhance the RL agents' learning efficiency is to leverage optimized topologies for pretraining. This approach involves a two-step training pipeline: starting with an imitation learning phase where the policy network is pretrained using supervised behavioral cloning [33] or adversarial imitation learning [34] from optimal topologies obtained via conventional optimization, followed by a pure policy training phase through RL. Initializing the policy with expert data can provide the RL agents with a strong starting point, rather than beginning from a random policy, similar to providing students with basic TO theory. This method has proven effective in complex environments such as StarCraft II [35] and the game of Go [36], leading to significant breakthroughs in those challenging domains.

# Conclusion

This paper introduces the Structural Optimization gym (SOgym), a novel RL environment designed for advancing the application of machine learning in TO. By framing the TO objective as an RL task, SOgym enables the development and training of RL agents to explore and exploit design spaces for optimal structural solutions.

The baseline results illustrate progress in training RL agents to solve continuous TO problems. This suggests the potential of RL, especially model-based algorithms like DreamerV3, in producing structurally robust designs. Despite beginning from a random policy, these agents successfully learned to produce connected topologies that were within 54% of the compliance achieved through conventional optimization. The comparison between model-free (PPO) and model-based (DreamerV3) approaches highlights the significant benefits of incorporating planning and world models, especially in enhancing sample efficiency and performance in computationally intensive environments like SOgym.

However, it is important to note that the primary goal of this paper is to introduce the SOgym environment and present initial baseline results, providing a preliminary understanding of what can be achieved with this new tool. These results are intended as a starting point rather than a definitive solution, and further research is needed to fully explore and validate the capabilities of RL in structural optimization.

Additionally, RL's framework presents opportunities for incorporating qualitative criteria into the optimization process. Inspired by the success of Reinforcement Learning from Human Feedback (RLHF) in fine-tuning large language models based on human preferences [37], future research could explore similar methods in structural optimization. By integrating qualitative aspects such as aesthetics into the reward function, RL agents could potentially be trained to produce designs that balance both functional and visual appeal. This approach could lead to innovative solutions that not only meet technical specifications but also enhance architectural beauty and user satisfaction.

Beyond exploring different environment objectives, the SOgym environment can be extended to support three-dimensional structures. This expansion would enable the exploration of more complex and realistic design scenarios, broadening the applicability of RL in structural engineering.

In summary, RL presents an interesting take on building up intuition for minimum compliance TO problems using ML. Moreover, the SOgym provides a valuable foundation for ongoing and future research and paves the way for future human feedback interaction which could prove a valuable design tool.

# Data & Code Availability

The SOgym environment is open-source and available at https://github.com/ThomasRochefortB/sogym

# Usage of Generative AI

During the preparation of this work, the authors utilized GPT-4o (version 2024-05-13) to enhance the clarity and conciseness of the language of their own text. After using this tool/service, the authors also reviewed and edited the AI-generated results as needed and they take full responsibility for the content of the paper.

# Acknowledgment

We acknowledge the support of the Natural Sciences and Engineering Research Council of Canada (NSERC) [funding reference number 569251]. This research was enabled in part by support provided by Calcul Québec (https://www.calculquebec.ca) and the Digital Research Alliance of Canada (DRAC) (alliancecan.ca).

# References

[1] M. P. Bendsøe and N. Kikuchi, "Generating optimal topologies in structural design using a homogenization method," *Computer methods in applied mechanics and engineering,* vol. 71, no. 2, pp. 197-224, 1988.

[2] M. P. Bendsøe and O. Sigmund, "Material interpolation schemes in topology optimization," *Archive of Applied Mechanics,* vol. 69, no. 9, pp. 635-654, 1999/11/01 1999, doi: 10.1007/s004190050248.

[3] E. Andreassen, A. Clausen, M. Schevenels, B. S. Lazarov, and O. Sigmund, "Efficient topology optimization in MATLAB using 88 lines of code," *Structural and Multidisciplinary Optimization,* vol. 43, no. 1, pp. 1-16, 2011/01/01 2011, doi: 10.1007/s00158-010-0594-7.

[4] L. Meng *et al.*, "From Topology Optimization Design to Additive Manufacturing: Today's Success and Tomorrow's Roadmap," *Archives of Computational Methods in Engineering,* vol. 27, no. 3, pp. 805-830, 2020/07/01 2020, doi: 10.1007/s11831-019-09331-1.

[5] J.-F. Gamache, A. Vadean, M. Capo, T. Rochefort-Beaudoin, N. Dodane, and S. Achiche, "Complexity-driven layout exploration for aircraft structures," *Design Science,* vol. 9, p. e13, 2023, Art no. e13, doi: 10.1017/dsj.2023.12.

[6] M. Baandrup, O. Sigmund, H. Polk, and N. Aage, "Closing the gap towards super-long suspension bridges using computational morphogenesis," *Nature Communications,* vol. 11, no. 1, p. 2735, 2020/06/01 2020, doi: 10.1038/s41467-020-16599-6.

[7] N. Aage, E. Andreassen, B. S. Lazarov, and O. Sigmund, "Giga-voxel computational morphogenesis for structural design," *Nature,* vol. 550, no. 7674, pp. 84-86, 2017/10/01 2017, doi: 10.1038/nature23911.

[8] J. Liu *et al.*, "Current and future trends in topology optimization for additive manufacturing," *Structural and Multidisciplinary Optimization,* vol. 57, no. 6, pp. 2457-2483, 2018/06/01 2018, doi: 10.1007/s00158-018-1994-3.

[9] L. Xia and P. Breitkopf, "Recent Advances on Topology Optimization of Multiscale Nonlinear Structures," *Archives of Computational Methods in Engineering,* vol. 24, no. 2, pp. 227-249, 2017/04/01 2017, doi: 10.1007/s11831-016-9170-7.

[10] S. Mukherjee *et al.*, "Accelerating Large-scale Topology Optimization: State-of-the-Art and Challenges," *Archives of Computational Methods in Engineering,* vol. 28, no. 7, pp. 4549-4571, 2021/12/01 2021, doi: 10.1007/s11831-021-09544-3.

[11] T. Rochefort-Beaudoin, A. Vadean, J.-F. Gamache, and S. Achiche, "Supervised deep learning for the moving morphable components topology optimization framework," *Engineering Applications of Artificial Intelligence,* vol. 123, p. 106436, 2023/08/01/ 2023, doi: https://doi.org/10.1016/j.engappai.2023.106436.

[12] R. V. Woldseth, N. Aage, J. A. Bærentzen, and O. Sigmund, "On the use of Artificial Neural Networks in Topology Optimisation," p. arXiv:2208.02563. [Online]. Available: https://ui.adsabs.harvard.edu/abs/2022arXiv220802563W

[13] S. Hoyer, J. Sohl-Dickstein, and S. Greydanus, "Neural reparameterization improves structural optimization," *arXiv preprint arXiv:1909.04240,* 2019.

[14] A. Chandrasekhar and K. Suresh, "TOuNN: Topology Optimization using Neural Networks," *Structural and Multidisciplinary Optimization,* vol. 63, no. 3, pp. 1135-1149, 2021/03/01 2021, doi: 10.1007/s00158-020-02748-4.

[15] M. Nobel-Jørgensen, D. Malmgren-Hansen, J. A. Bærentzen, O. Sigmund, and N. Aage, "Improving topology optimization intuition through games," *Structural and Multidisciplinary Optimization,* vol. 54, no. 4, pp. 775-781, 2016/10/01 2016, doi: 10.1007/s00158-016-1443-0.

[16] K. Hayashi and M. Ohsaki, "Reinforcement Learning and Graph Embedding for Binary Truss Topology Optimization Under Stress and Displacement Constraints," (in English), *Frontiers in Built Environment,* Original Research vol. 6, 2020-April-30 2020, doi: 10.3389/fbuil.2020.00059.

[17] N. K. Brown, A. P. Garland, G. M. Fadel, and G. Li, "Deep reinforcement learning for engineering design through topology optimization of elementally discretized design domains," *Materials & Design,* vol. 218, p. 110672, 2022.

[18] G. Brockman *et al.*, "Openai gym," *arXiv preprint arXiv:1606.01540,* 2016.

[19] F. Wein, P. D. Dunning, and J. A. Norato, "A review on feature-mapping methods for structural optimization," *Structural and Multidisciplinary Optimization,* vol. 62, no. 4, pp. 1597-1638, 2020/10/01 2020, doi: 10.1007/s00158-020-02649-6.

[20] W. Zhang, J. Yuan, J. Zhang, and X. Guo, "A new topology optimization approach based on Moving Morphable Components (MMC) and the ersatz material model," *Structural and Multidisciplinary Optimization,* vol. 53, no. 6, pp. 1243-1260, 2016/06/01 2016, doi: 10.1007/s00158-015-1372-3.

[21] D. Hafner, J. Pasukonis, J. Ba, and T. Lillicrap, "Mastering diverse domains through world models," *arXiv preprint arXiv:2301.04104,* 2023.

[22] J. Schulman, F. Wolski, P. Dhariwal, A. Radford, and O. Klimov, "Proximal policy optimization algorithms," *arXiv preprint arXiv:1707.06347,* 2017.

[23] A. Mirhoseini *et al.*, "A graph placement methodology for fast chip design," *Nature,* vol. 594, no. 7862, pp. 207-212, 2021/06/01 2021, doi: 10.1038/s41586-021-03544-w.

[24] C. Berner *et al.*, "Dota 2 with large scale deep reinforcement learning," *arXiv preprint arXiv:1912.06680,* 2019.

[25] L. Ouyang *et al.*, "Training language models to follow instructions with human feedback," *Advances in Neural Information Processing Systems,* vol. 35, pp. 27730-27744, 2022.

[26] L. Espeholt *et al.*, "Impala: Scalable distributed deep-rl with importance weighted actor-learner architectures," in *International conference on machine learning*, 2018: PMLR, pp. 1407-1416.

[27] K. Cobbe, C. Hesse, J. Hilton, and J. Schulman, "Leveraging procedural generation to benchmark reinforcement learning," in *International conference on machine learning*, 2020: PMLR, pp. 2048-2056.

[28] M. Andrychowicz *et al.*, "What matters in on-policy reinforcement learning? a large-scale empirical study," *arXiv preprint arXiv:2006.05990,* 2020.

[29] K. Svanberg, "MMA and GCMMA: two methods for nonlinear optimization," 2014.

[30] T. Rochefort-Beaudoin, A. Vadean, J.-F. Gamache, and S. Achiche, "Comparative Study of First-Order Moving Asymptotes Optimizers for the Moving Morphable Components Topology Optimization Framework," in *ASME 2022 International Design Engineering Technical Conferences and Computers and Information in Engineering Conference*, 2022, vol. Volume 2: 42nd Computers and Information in Engineering Conference (CIE), V002T02A028, doi: 10.1115/detc2022-88722. [Online]. Available: https://doi.org/10.1115/DETC2022-88722

[31] V. Keshavarzzadeh, M. Alirezaei, T. Tasdizen, and R. M. Kirby, "Image-Based Multiresolution Topology Optimization Using Deep Disjunctive Normal Shape Model," *Computer-Aided Design,* vol. 130, p. 102947, 2021/01/01/ 2021, doi: https://doi.org/10.1016/j.cad.2020.102947.

[32] P. Henderson, R. Islam, P. Bachman, J. Pineau, D. Precup, and D. Meger, "Deep reinforcement learning that matters," in *Proceedings of the AAAI conference on artificial intelligence*, 2018, vol. 32, no. 1.

[33] P. Abbeel and A. Y. Ng, "Apprenticeship learning via inverse reinforcement learning," in *Proceedings of the twenty-first international conference on Machine learning*, 2004, p. 1.

[34] J. Ho and S. Ermon, "Generative adversarial imitation learning," *Advances in neural information processing systems,* vol. 29, 2016.

[35] O. Vinyals *et al.*, "Grandmaster level in StarCraft II using multi-agent reinforcement learning," *Nature,* vol. 575, no. 7782, pp. 350-354, 2019/11/01 2019, doi: 10.1038/s41586-019-1724-z.

[36] D. Silver *et al.*, "Mastering the game of Go with deep neural networks and tree search," *Nature,* vol. 529, no. 7587, pp. 484-489, 2016/01/01 2016, doi: 10.1038/nature16961.

[37] P. F. Christiano, J. Leike, T. Brown, M. Martic, S. Legg, and D. Amodei, "Deep reinforcement learning from human preferences," *Advances in neural information processing systems,* vol. 30, 2017.

# Appendix A – Hybrid Optimizer Parameters

Conventional optimization was run using a hybrid MMA-GCMMA optimizer which switched from MMA to GCMMA after detecting oscillations in the objective function according to a *switch* criterion. A maximum of 1000 outer iterations were allowed during the iterative process. A low resolution of 50 elements per dimension unit was used to fit the resolution of SOgym. Table 7 describes the parameters defining the MMA and GCMMA optimizer.

Table 7: Parameters used for the hybrid MMA-GCMMA optimizer.

| Solver | Parameter | Description | Value |
|---|---|---|---|
| **MMA / GCMMA** | *epsimin* | Termination value of $\varepsilon$ | $10^{-10}$ |
| | *raa0* | Coefficient in the update of $p_{ij}^{(k,v)}$ and $q_{ij}^{(k,v)}$ | 0.01 |
| | *albefa* | Coefficient for the update of the design variables bounds $\alpha_j^k$ and $\beta_j^k$. | 0.4 |
| | *asyinit* | Initial step size of lower and upper asymptote update | 0.05 |
| | *asyincr* | Increasing lower and upper asymptote step size | 0.8 |
| | *asydecr* | Decreasing lower and upper asymptote step size. | 0.6 |
| | *c* | Column vector with the constants $c_i$ in the terms $c_i y_i$ | 1000 |
| | *d* | Column vector with the constants $d_i$ in the terms $\frac{1}{2} d_i y_i^2$ | 1 |
| | $a_0$ | Constant in the term $a_0 z$ | 1 |
| | *a* | Column vector with the constants $a_i$ in the terms $a_i z$ | 0 |
| **GCMMA** | *maxinnerit* | Maximum number of inner iterations for the GCMMA optimizer | 2 |
| | *move* | Coefficient for the update of the design variables bounds $\alpha_j^k$ and $\beta_j^k$. | 1.0 |
| **Hybrid** | *switch* | Criterion to switch between MMA and GCMMA in the hybrid optimizer | $2e^{-5}$ |